\documentclass[nohyperref]{article}

\usepackage{microtype}
\usepackage{graphicx}
\usepackage{subfigure}
\usepackage{booktabs} %

\usepackage{hyperref}

\usepackage[accepted]{icml2022modified}

\usepackage{amsmath}
\usepackage{amssymb}
\usepackage{mathtools}
\usepackage{amsthm}

\usepackage[capitalize,noabbrev]{cleveref}

\usepackage{amsmath}
\usepackage{amssymb}
\usepackage{xspace}
\usepackage{graphicx}
\usepackage{array}
\usepackage{enumitem}
\usepackage{tikz}
\usepackage{tikz-network}
\usepackage{adjustbox}

\usepackage{algorithm}
\usepackage{algorithmic}

\usepackage{wrapfig}

\usepackage{booktabs}       %

\DeclareMathOperator*{\argmax}{\arg\!\max}
\DeclareMathOperator*{\argmin}{\arg\!\min}

\theoremstyle{plain}

\theoremstyle{definition}

\theoremstyle{remark}

\usepackage[textsize=tiny]{todonotes}

\icmltitlerunning{Planning Spatial Networks with Monte Carlo Tree Search}

\begin{document}

\twocolumn[
\icmltitle{Planning Spatial Networks with Monte Carlo Tree Search}

\icmlsetsymbol{equal}{*}

\begin{icmlauthorlist}
\icmlauthor{Victor-Alexandru Darvariu}{ucl,ati}
\icmlauthor{Stephen Hailes}{ucl}
\icmlauthor{Mirco Musolesi}{ucl,ati,blq}
\end{icmlauthorlist}

\icmlaffiliation{ucl}{Department of Computer Science, University College London, London, United Kingdom}
\icmlaffiliation{ati}{The Alan Turing Institute, London, United Kingdom}
\icmlaffiliation{blq}{Department of Computer Science and Engineering, University of Bologna, Bologna, Italy}

\icmlcorrespondingauthor{Victor-Alexandru Darvariu}{v.darvariu@ucl.ac.uk}

\icmlkeywords{Monte Carlo Tree Search, Graph Neural Networks, Spatial Networks, Network Science}

\vskip 0.3in
]

\printAffiliationsAndNotice{}  %

\begin{abstract}
We tackle the problem of goal-directed graph construction: given a starting graph, a budget of modifications, and a global objective function, the aim is to find a set of edges whose addition to the graph achieves the maximum improvement in the objective (e.g., communication efficiency). This problem emerges in many networks of great importance for society such as transportation and critical infrastructure networks. We identify two significant shortcomings with present methods. Firstly, they focus exclusively on network topology while ignoring spatial information; however, in many real-world networks, nodes are embedded in space, which yields different global objectives and governs the range and density of realizable connections. Secondly, existing RL methods scale poorly to large networks due to the high cost of training a model and the scaling factors of the action space and global objectives. 

In this work, we formulate this problem as a deterministic MDP. We adopt the Monte Carlo Tree Search framework for planning in this domain, prioritizing the optimality of final solutions over the speed of policy evaluation. We propose several improvements over the standard UCT algorithm for this family of problems, addressing their single-agent nature, the trade-off between the costs of edges and their contribution to the objective, and an action space linear in the number of nodes. We demonstrate the suitability of this approach for improving the global efficiency and attack resilience of a variety of synthetic and real-world networks, including Internet backbone networks and metro systems. Our approach obtains a 24\% improvement in these metrics compared to UCT on the largest networks tested and scalability superior to previous methods.
\end{abstract}

\section{Introduction}\label{intro}
Graphs are a pervasive representation that arises naturally in a variety of disciplines; however, their non-Euclidean structure has traditionally proven challenging for machine learning and decision-making approaches. The emergence of the Graph Neural Network learning paradigm~\cite{scarselli_graph_2009} and geometric deep learning more broadly~\cite{bronstein_geometric_2017} have brought about encouraging breakthroughs in diverse application areas for graph-structured data: relevant examples include combinatorial optimization~\cite{vinyals_pointer_2015,bello_neural_2016,khalil_learning_2017}, recommendation systems~\cite{monti_geometricmatrix_2017,ying_graphconvolutional_2018} and computational chemistry~\cite{gilmer_neural_2017,jin_junction_2019,you_graph_2018,bradshawModelSearchSynthesizable2019}.

\definecolor{myorange}{RGB}{238, 122, 0}
\definecolor{myred}{RGB}{134, 0, 0}
\definecolor{myblue}{RGB}{0, 0, 204}
\definecolor{mypurple}{RGB}{76, 0, 153}

\begin{figure*}[t]
\begin{center}
\includegraphics[width=0.85\textwidth]{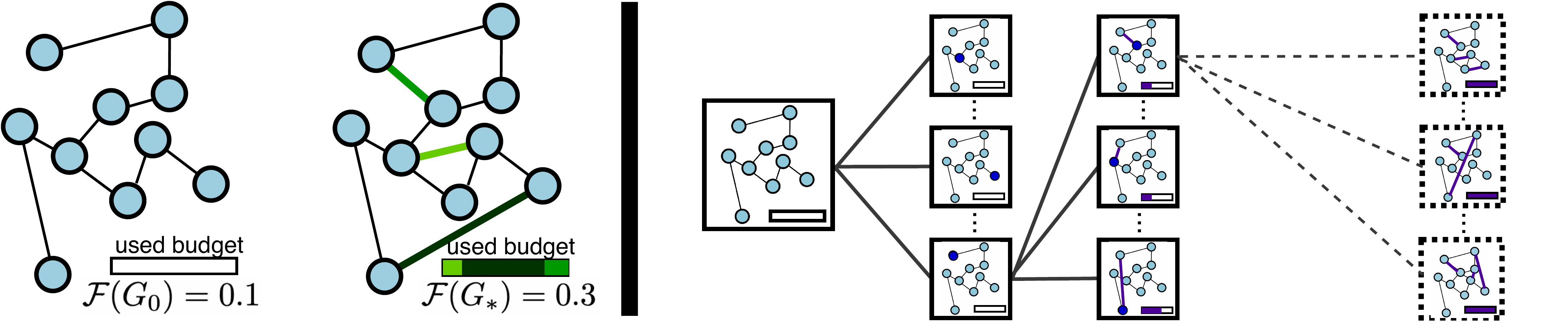} 
\caption{Schematic of our approach. Left: given a spatial graph $G_0$, an objective function $\mathcal{F}$, and a budget defined in terms of edge lengths, the goal is to add a set of edges such that the resulting graph $G_*$ maximally increases $\mathcal{F}$. Right: we formulate this problem as a deterministic MDP, in which states are graphs, \textcolor{myblue}{actions} represent the selection of a node, transitions add an \textcolor{mypurple}{edge} every two steps, and the reward is based on $\mathcal{F}$. We use Monte Carlo Tree Search to plan \textit{the optimal set of edges to be added} using knowledge of this MDP, and propose a method (SG-UCT) that improves on standard UCT.
} 
\label{psn_illustration}
\end{center}
\end{figure*}

There is an increasing interest in the problem goal-directed graph construction, in which the aim is to build or modify the topology of a graph (i.e., add a set of edges) so as to maximize the value of a global objective function. Unlike classic graph algorithms, which assume that the graph topology is \textit{static}, in this setting the graph structure itself is \textit{dynamically} changed. As this task involves an element of exploration (optimal solutions are not known a priori), its formulation as a decision-making process is a suitable paradigm. Model-free reinforcement learning (RL) techniques have been applied in the context of the derivation of adversarial examples for graph-based classifiers~\cite{dai_adversarial_2018} and generation of molecular graphs~\cite{you_graph_2018}.~\citet{darvariu2021goal} have formulated the optimization of a global structural graph property as an MDP and approached it using a variant of the RL-S2V~\cite{dai_adversarial_2018} algorithm, showing that generalizable strategies for improving a global network objective can be learned, and can obtain performance superior to prior approaches~\cite{beygelzimer_improving_2005,schneider_mitigation_2011,wangAlgebraicConnectivityOptimization2008,wang_improving_2014} in some cases. However, when applying such approaches to improve the properties of real-world networked systems, two challenges become apparent:

\begin{enumerate}
	\item \textit{Inability to account for spatial properties of the graphs}: optimizing the topology of the graph alone is only part of the problem in many cases. A variety of real-world networks share the property that nodes are embedded in space, and this geometry has a strong relationship with the types of topologies that can be created~\cite{gastner_spatialstructure_2006,barthelemySpatialNetworks2011}. Since there is a cost associated with edge length, connections tend to be local, and long-range connections must be justified by some gain (e.g., providing connectivity to a hub). Furthermore, objective functions defined over nodes' positions (such as efficiency) are key for understanding their organization~\cite{latoraEfficientBehaviorSmallWorld2001}. 
	\item \textit{Scalability}: existing methods based on RL are challenging to scale, due to the sample complexity of current training algorithms, the linear increase of possible actions in the number of nodes, and the complexity of evaluating the global objectives (typically polynomial in the number of nodes). Additionally, training data (i.e., instances of real-world graphs) are scarce and we are typically interested in a specific starting graph (e.g., a particular infrastructure network to be improved).
\end{enumerate}

In this paper, we set out to address these shortcomings. For the first time, we consider the problem of the construction of spatial graphs as a decision-making process that explicitly captures the influence of space on graph-level objectives, realizable links between nodes, and connection budgets. Furthermore, to address the scalability issue, we propose to use planning methods in order to plan an \textit{optimal set of edges} to add to the graph, which sidesteps the problem of sample complexity since we do not need to learn a policy. We adopt the Monte Carlo Tree Search framework -- specifically, the UCT algorithm~\cite{kocsis_bandit_2006} -- and show it can successfully be applied in planning graph construction strategies. We illustrate our approach at a high level in Figure~\ref{psn_illustration}. Finally, we propose several improvements over the basic UCT method in the context of spatial networks. These relate to important characteristics of this family of problems: namely, their single-agent, deterministic nature; the inherent trade-off between the cost of edges and their contribution to the global objective; and an action  space that is linear in the number of nodes in the network. Our proposed approach, Spatial Graph UCT (SG-UCT), is designed with these characteristics in mind and it relies on a very limited set of assumptions. For this reason, it can be applied to a vast class of spatial networked systems.

As objective functions, in this study, we consider the global properties of network efficiency and robustness to targeted attacks. While these represent a variety of practical scenarios, our approach is broadly applicable to any other structural property. We perform an evaluation on synthetic graphs (generated by a spatial growth model) and several real-world graphs (internet backbone networks and metro transportation networks), comparing SG-UCT to UCT as well as a variety of baselines that have been proposed in the past. Our results show that SG-UCT performs best out of all methods in all the settings we tested; moreover, the performance gain over UCT is substantial (24\% on average and up to 54\% over UCT on the largest networks tested in terms of a robustness metric). In addition, we conduct an ablation study that explores the impact of the algorithmic components.
\section{Preliminaries}\label{prelims}

\noindent \textbf{MDPs and Planning.} Markov Decision Processes (MDPs) are widely adopted for effective formalization of decision making tasks. The decision maker, usually called an \textit{agent}, interacts with an \textit{environment}.  When in a \textit{state} $s\in \mathcal{S}$, the agent must take an \textit{action} $a$ out of the set $\mathcal{A}(s)$ of valid actions, receiving a \textit{reward} $r$ governed by the reward function $\mathcal{R}(s,a)$. Finally, the agent finds itself in a new state $s'$, depending on a transition model $\mathcal{P}$ that governs the joint probability distribution $P(s',a,s)$ of transitioning to state $s'$ after taking action $a$ in state $s$. This sequence of interactions gives rise to a \textit{trajectory} $S_0, A_0, R_1, S_1, A_1, R_2, \ldots S_T$, which continues until a terminal state $S_T$ is reached. In \textit{deterministic} MDPs, there exists a unique state $s'$ s.t. $P(S_{t+1}=s'|S_t=s,A_t=a)=1$. The tuple $(\mathcal{S,A,P,R, \gamma})$ defines this MDP, where $\gamma \in [0,1]$ is a discount factor. We also define a \textit{policy} $\pi(a|s)$, a distribution of actions over states. There exists a spectrum of algorithms for constructing a policy, ranging from model-based algorithms (which assume knowledge of the MDP) to model-free algorithms (which require only samples of agent-environment interactions). In the cases in which either the full MDP specification or a model are available, \textit{planning} can be used to construct a policy, for example by using forward search~\cite{russell2010artificial}.%

\noindent \textbf{Monte Carlo Tree Search.} Monte Carlo Tree Search (MCTS) is a model-based planning technique that addresses the inability to explore all paths in large MDPs by constructing a policy \textit{from the current state}. Values of states are estimated through the returns obtained by executing simulations from the starting state. Upper Confidence Bounds for Trees (UCT), a variant of MCTS, consists of a tree search in which the decision at each node is framed as an independent multi-armed bandit problem. At decision time, the \textit{tree policy} of the algorithm selects the child node corresponding to action $a$ that maximizes $UCT(s,a) = \frac{R(s,a)}{N(s,a)} + 2  c_p  \sqrt{\frac{2  \ln{N(s)}}{N(s,a)}}$, where $R(s,a)$ is the sum of returns obtained when taking action $a$ in state $s$, $N(s)$ is the number of parent node visits, $N(s,a)$ the number of child node visits, and $c_p$ is a constant that controls the level of exploration \cite{kocsis_bandit_2006}. In the standard version of the algorithm, the returns are estimated using a random \textit{default policy} when expanding a node. MCTS has been applied to great success in connectionist games such as Morpion Solitaire~\cite{rosin2011nested}, Hex~\cite{nash_games_1952,anthony_thinking_2017}, and Go, which was previously thought computationally intractable~\cite{silver_alphago_2016,silver_general_2018}. 

\noindent \textbf{Spatial Networks.} We define a \textit{spatial network} as the tuple $G=(V,E,f,w)$. $V$ is the set of vertices, and $E$ is the set of edges. $f \colon V \to M$ is a function that maps nodes in the graph to a set of positions $M$. We require that $M$ admits a metric $d$, i.e., there exists a function $d \colon M \times M \to \mathbb{R}^{+}$ defining a pairwise \textit{distance} between elements in $M$. The tuple $(M,d)$ defines a \textit{space}, common examples of which include Euclidean space and spherical geometry. $w \colon E \to \mathbb{R}^{+}$ associates a \textit{weight} with each edge: a positive real-valued number that denotes its capacity.

\noindent \textbf{Global Objectives in Spatial Networks.} We consider two global objectives $\mathcal{F}$ for spatial networks that are representative of a wide class of properties relevant in real-world situations. Depending on the domain, there are many other global objectives for spatial networks that can be considered, to which the approach that we present is directly applicable. 

\textit{Efficiency.} Efficiency is a metric quantifying how well a network exchanges information. It measures how fast information can travel between any pair of nodes in the network on average, and is hypothesized to be an underlying principle for the organization of networks~\cite{latoraEfficientBehaviorSmallWorld2001}. Efficiency does not solely depend on topology but also on the \textit{spatial} distances between the nodes in the network. We adopt the definition of global efficiency as formalized by~\citeauthor{latoraEfficientBehaviorSmallWorld2001}, and let $\mathcal{F}_E(G) =  \frac{1}{N(N-1)} \sum_{i \neq j \in V}{\frac{1}{ sp(i,j) }}$, where $sp(i,j)$ is the cumulative length of the shortest path between vertices $i$ and $j$. To normalize, we divide by the ideal efficiency $\mathcal{F}_E^*(G) =  \frac{1}{N(N-1)} \sum_{i \neq j \in V}{\frac{1}{ d(i,j) }}$, and possible values are thus in $[0,1]$.\footnote{It is worth noting that efficiency is a more suitable metric for measuring the exchange of information than the inverse average path length between pairs of nodes. In the extreme case where the network is disconnected (and thus some paths lengths are infinite), this metric does not go to infinity. More generally, this metric is better suited for systems in which information is exchanged in a parallel, rather than sequential, way~\cite{latoraEfficientBehaviorSmallWorld2001}.}
Efficiency is computable in $O(|V|^3)$ by using e.g. the Floyd-Warshall shortest path algorithm.\footnote{In practice, this may be made faster by considering dynamic shortest path algorithms, e.g.~\cite{demetrescu2004new}.}

\textit{Robustness.} We consider the property of robustness, i.e., the resilience of the network in the face of removals of nodes. 
We adopt a robustness measure widely used in the literature~\cite{albert_error_2000, PhysRevLett.85.5468} and of practical interest and applicability based on the largest connected component (LCC), i.e., the component with most nodes. In particular, we use the definition in~\cite{schneider_mitigation_2011}, which considers the size of the LCC as nodes are removed from the network. We consider only the targeted attack case as previous work has found it is more challenging~\cite{albert_error_2000,darvariu2021goal}. We define the robustness measure as $\mathcal{F}_R(G) =  \mathbb{E}_{\xi}[\frac{1}{N} \sum_{i=1}^{N}{s(G, \xi, i)} ]$, where $s(G, \xi, i)$ denotes the fraction of nodes in the LCC of $G$ after the removal of the first $i$ nodes in the permutation $\xi$ (in which nodes appear in descending order of their degrees). Possible values are in $[\frac{1}{N},0.5)$. This quantity can be estimated using Monte Carlo simulations and scales as $O(|V|^2 \times (|V| + |E|))$.

It is worth noting that the value of the objective functions typically increases the more edges exist in the network (the complete graph has both the highest possible efficiency and robustness). However, constructing a complete graph is wasteful, and so it may be necessary to balance the contribution of an edge to the objective with its cost. The proposed method explicitly accounts for this trade-off, which is widely observed in infrastructure and brain networks~\cite{gastner_shapeefficiency_2006,bullmoreEconomyBrainNetwork2012}.

\section{Proposed Method}\label{method}

In this section, we first formulate the construction of spatial networks in terms of a global objective function as an MDP. Subsequently, we propose a variant of the UCT algorithm (SG-UCT) for planning in this MDP, which exploits the characteristics of spatial networks.

\subsection{Spatial Graph Construction as an MDP}

\noindent \textbf{Spatial Constraints in Network Construction.} Spatial networks that can be observed in the real world typically incur a cost to edge creation. Take the example of a power grid: the cost of a link can be expressed as a function of its geographic distance as well as its capacity. It is vital to consider both aspects of link cost in the process of network construction. We let $c(i,j)$ denote the cost of edge $(i,j)$ and $C(\Gamma)=\sum_{(i,j) \in \Gamma}{c(i,j)}$ be the cost of a \textit{set} of edges $\Gamma$. We consider $c(i,j)=w(i,j)*d(f(i),f(j))$ to capture the notion that longer, higher capacity connections are more expensive -- although different notions of cost may be desirable depending on the domain. To ensure fair comparisons, we normalize costs $c(i,j)$ to be in $[0,1]$.

\noindent \textbf{Problem Statement.} Let $\mathbf{G}^{(N)}$ be the set of labeled, undirected, weighted, spatial networks with $N$ nodes. We let $\mathcal{F}\colon \mathbf{G}^{(N)} \to [0,1]$ be an objective function, and $b_0\in\mathbb{R}^{+}$ be a modification budget. Given an initial graph $G_0=(V,E_0,f,w)\in\mathbf{G}^{(N)}$, the aim is to add a set of edges $\Gamma$ to $G_0$ such that the graph $G_*=(V,E_*,f,w)$ satisfies: 

\begin{align}
    G_*&=\argmax_{G' \in \mathbf{G}'} {\mathcal{F}(G')}, \label{eq:1} \\
    \text{where}\quad \mathbf{G}' &= \{ G \in \mathbf{G}^{(N)}\ |\ E = E_0 \cup \Gamma \ . \ C(\Gamma) \leq b_0 \} \nonumber 
\end{align}

\noindent \textbf{MDP Formulation.} We next define the MDP elements:

\textit{State}: The state $S_t$ is a 3-tuple $\left(G_t,\sigma_t,b_t\right)$ containing the spatial graph $G_t=(V, E_t,f,w)$, an \textit{edge stub} $\sigma_t \in V$, and the remaining budget $b_t$. $\sigma_t$ can be either the empty set $\varnothing$ or the singleton $\{v\}$, where $v \in V$. If the edge stub is non-empty, it means that the agent has ``committed" in the previous step to creating an edge originating at the edge stub.

\textit{Action}: For scalability to large graphs, an action $A_t$ corresponds to the selection of a \textit{single} node in $V$ (thus having at most $|V|$ choices). We enforce spatial constraints as follows: given a node $i$, we define the set $\mathcal{K}(i)$ of \textit{connectable} nodes $j$ that represent realizable connections. We let:
\[
\mathcal{K}(i) = \{j \in V\ |\ c(i,j) \leq \rho \max_{k \in V . (i,k) \in E_0}{c(i,k)} \}
\]
which formalizes the idea that a node can only connect as far as a proportion $\rho$ of its longest existing connection, with $\mathcal{K}(i)$ fixed based on the initial graph $G_0$. This has the benefit of allowing long-range connections if they already exist in the network. Given an unspent connection budget $b_t$, we let the set $\mathcal{B}(i,b_t) = \{ j \in \mathcal{K}(i)\ |\ c(i,j) \leq b_t \}$ consist of those connectable nodes whose cost is not more than the unspent budget. Letting the degree of node $v$ be $d_v$, available actions are defined as:\footnote{Depending on the type of network being considered, in practice there may be different types of constraints on the connections that can be realized. For example, in transportation networks, there can be obstacles that make link creation impossible, such as prohibitive landforms or populated areas. In circuits and utility lines, planarity is a desirable characteristic as it makes circuit design cheaper. Such constraints can be captured by the definition of $\mathcal{K}(i)$ and enforced by the environment when providing the agent with available actions $\mathcal{A}(s)$. Conversely, defining $\mathcal{K}(i) = V \setminus \{i\}$ recovers the simplified case where no constraints are imposed.}
\begin{align*}
\mathcal{A}(S_t&=((V, E_t, f, w), \varnothing, b_t)) \\ 
&= \{v \in V\ |\ d_v < |V| - 1 \ \land\ |\mathcal{B}(v, b_t)| > 0 \} \\
\mathcal{A}(S_t&=((V, E_t, f, w), \{ \sigma_t \}, b_t)) \\
&= \{v \in V\ |\ (\sigma_t, v) \notin E_t\ \land\ v \in \mathcal{B}(\sigma_t, b_t) \}
\end{align*}

\textit{Transitions}: The deterministic transition model adds an edge every two steps. Concretely, we define it as $P(S_t=s'|S_{t-1}=s,A_{t-1}=a) = \delta_{S_ts'}$, where $s^{\prime}=$
\begin{align*}
 &((V,E_{t-1} \cup (\sigma_{t-1}, a ), f, w), \varnothing, b_{t-1} - c(\sigma_{t-1}, a) ),\  \text{if}\  2\mid t \\
 &((V,E_{t-1}, f, w), \{a\}, b_{t-1}),\  \text{otherwise}
\end{align*}

\textit{Reward}: The final reward $R_{T-1}$ is defined as $\mathcal{F}(G_{T}) - \mathcal{F}(G_0)$ and all intermediary rewards are $0$.

Episodes in this MDP proceed for an arbitrary number of steps until the budget is exhausted and no valid actions remain (concretely, $|\mathcal{A}(S_t)| = 0$). Since we are in the finite horizon case, we let $\gamma=1$.
Given the MDP definition above, the problem specified in Equation~\ref{eq:1} can be reinterpreted as finding the trajectory ${\tau}_*$ that starts at $S_0=(G_0,\varnothing,b_0)$ such that the final reward $R_{T-1}$ is maximal -- actions along this trajectory will define the set of edges $\Gamma$.

\subsection{Algorithm}\label{algo}

The formulation above can, in principle, be used with any planning algorithm for MDPs in order to identify an optimal set of edges to add to the network. The UCT algorithm, discussed in Section~\ref{prelims}, is one such algorithm that has proven very effective in a variety of settings. We refer the reader to~\citet{browne_survey_2012} for an in-depth description of the algorithm and its various applications. However, the generic UCT algorithm assumes very little about the particulars of the problem under consideration, which, in the context of spatial network construction, may lead to sub-optimal solutions. In this section, we identify and address concerns specific to this family of problems, and formulate the Spatial Graph UCT (SG-UCT) variant of UCT in Algorithm~\ref{alg:sg-uct}. The evaluation presented in Section~\ref{experiments} compares SG-UCT to UCT and other baselines, and contains an ablation study of SG-UCT's components.

\noindent \textbf{Best Trajectory Memorization (BTM).} The standard UCT algorithm is applicable in a variety of settings, including multi-agent, stochastic environments. For example, in two-player games, an agent needs to re-plan from the new state that is arrived at after the opponent executes its move. However, the single-agent (puzzle), deterministic nature of the problem considered means that there is no-need to re-plan trajectories after a stochastic event: the agent can plan all its actions from the very beginning in a single step. We thus propose the following modification over UCT: memorizing the trajectory with the highest reward found during the rollouts, and returning it at the end of the search. We name this Best Trajectory Memorization, shortened BTM. This is similar in spirit (albeit much simpler) to ideas used in Reflexive and Nested MCTS for deterministic puzzles, where the best move found at lower levels of a nested search are used to inform the upper level~\cite{cazenave2007reflexive,cazenave2009nested}.

\noindent \textbf{Cost-Sensitive Default Policy.} The standard default policy used to perform out-of-tree actions in the UCT framework is based on random rollouts. While evaluating nodes using this approach is free from bias, rollouts can lead to high-variance estimates, which can hurt the performance of the search. Previous work has considered hand-crafted heuristics and learned policies as alternatives, although, perhaps counter-intuitively, learned policies may lead to worse results~\cite{gelly_combining_2007}. As initially discussed in Section~\ref{prelims}, the value of the objective functions we consider grows with the number of edges of the graph. We thus propose the following default policy for spatial networks: sampling each edge with probability inversely proportional to its cost. Formally, we let the probability of edge $i,j$ being selected during rollouts be proportional to $(\max_{i,j} {c(i,j)} - c(i,j))^\beta$, where $\beta$ denotes the level of bias. $\beta \to 0$ reduces to random choices, while $\beta \to \infty$ selects the minimum cost edge. This is very inexpensive computationally, as the edge costs only need to be calculated once, at the start of the search.

\noindent \textbf{Action Space Reduction.} In certain domains, the number of actions available to an agent is large, which can greatly affect scalability. Previous work in RL has considered decomposing actions into independent sub-actions~\cite{he_deepreinforcement_2016}, generalizing across similar actions by embedding them in a continuous space~\cite{dulac-arnold_deepreinforcement_2015}, or learning which actions to eliminate via supervision provided by the environment~\cite{zahavy_learnwhat_2018}. Existing approaches in planning consider progressively widening the search based on a heuristic~\cite{chaslot_progressivestrategies_2008} or learning a partial policy for eliminating actions in the search tree~\cite{pinto_learningpartial_2017}. 

\begin{figure}
   \begin{center}
   \includegraphics[width=0.50\columnwidth]{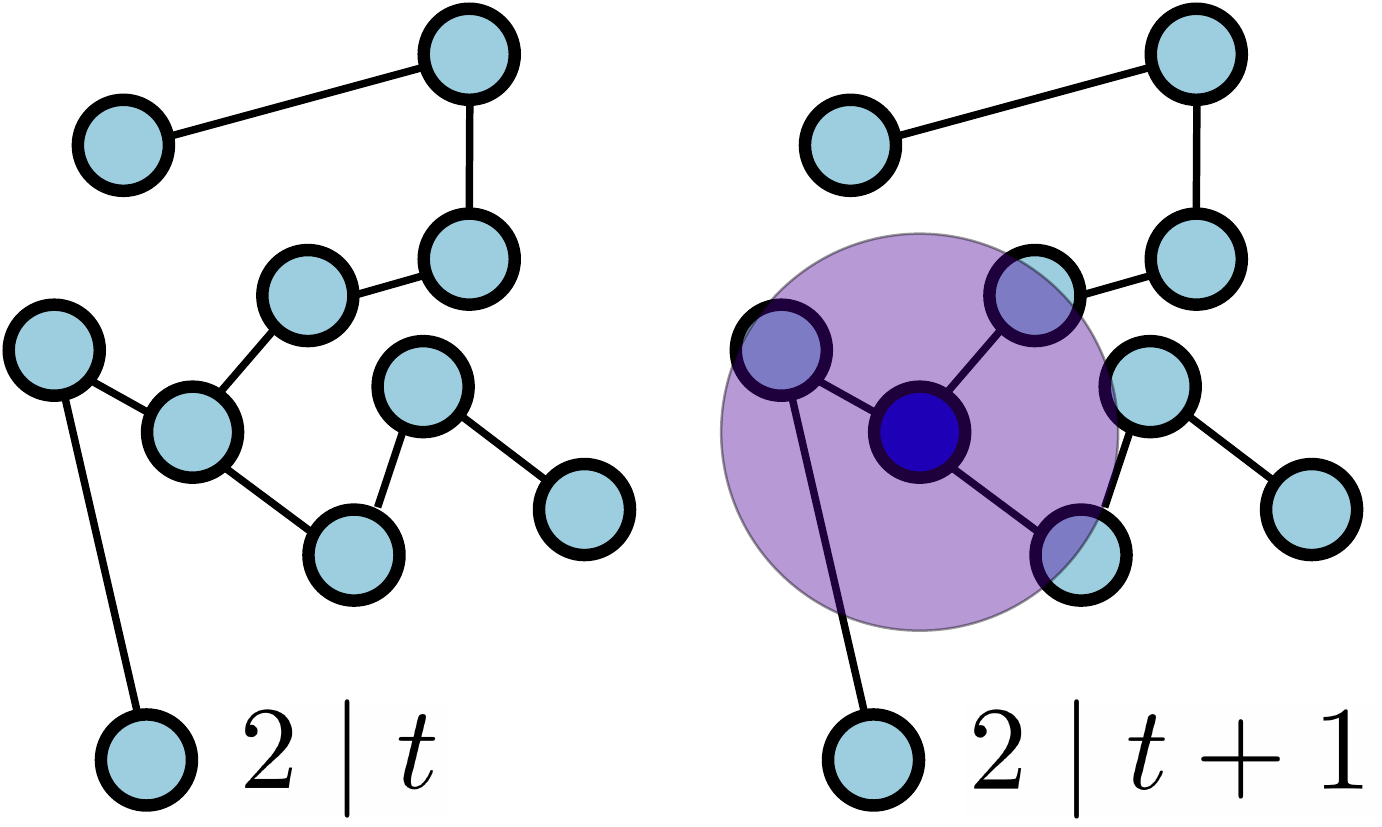}
   \caption{Illustration of the asymmetry in the number of actions at even (state has no edge stub) versus odd $t$ (state has an edge stub). Spatial constraints are imposed in the latter case, reducing the number of actions.}
   \label{tree-asymmetry}
   \end{center}
\end{figure}

Concretely, with the MDP definition used, the action space grows linearly in the number of nodes. This is partly addressed by the imposed connectivity constraints: once an edge stub is selected (equivalently, at odd values of $t$), the branching factor of the search is small since only \textit{connectable} nodes need to be considered. However, the number of actions when selecting the origin node of the edge (even values of $t$) remains large, which might become detrimental to performance as the size of the network grows (we illustrate this in Figure~\ref{tree-asymmetry}). Can this be mitigated? 

We consider limiting the nodes that can initiate connections to a specific subset -- which effectively prunes away all branches in the search tree that are not part of this set. Concretely, let a \textit{reduction policy} $\phi$ be a function that, given the initial graph $G_0$, outputs a strict subset of its nodes.\footnote{Learning a reduction policy in a data-driven way is also possible; however, obtaining the supervision signal (i.e., node rankings over multiple MCTS runs) is very expensive. Furthermore, since  we prioritize performance on specific graph instances over generalizable policies, simple statistics may be sufficient. Still, a learned reduction policy that predicts an entire set at once may be able to identify better subsets than individual statistics alone. Furthermore, a possible limitation of using a reduction policy is that the optimal trajectory in the full MDP may be excluded. We consider these worthwhile directions for future investigations.}
Then, we modify our definition of allowed actions as follows: under a reduction policy $\phi$, we define 
\begin{align*}
	\mathcal{A}_{\phi}(S_t) &= \mathcal{A}(S_t) \cap \phi(G_0),\  \text{if}\  2\mid t \\
							 &= \mathcal{A}(S_t),\   \text{otherwise}
\end{align*}

We investigate the following class of reduction policies: a node $i$ is included in $\phi(G_0)$ if and only if it is among the top nodes ranked by a local node statistic $\lambda(i)$. More specifically, we consider the $\lambda(i)$ listed below, where $\mathit{gain}(i,j)={\mathcal{F}(V,E \cup (i,j),f,w) - \mathcal{F}(V,E,f,w)}$. Since the performance of reduction strategies may depend heavily on $\mathcal{F}$, we treat it as a tunable hyperparameter.

\begin{itemize}[leftmargin=*]
\item \textit{Degree (DEG)}: $d_i$; \textit{Inverse Degree (ID)}: $\max_j {d_j} - d_i$; \textit{Number of Connections (NC)}: $|\mathcal{K}(i)|$
\item \textit{Best Edge (BE)}: $\max_{j \in \mathcal{K}(i) } \mathit{gain}(i,j)$; \textit{BE Cost Sensitive (BECS)}: $\max_{j \in \mathcal{K}(i) } {\frac{\mathit{gain}(i,j)}{c(i,j)}}$
\item \textit{Average Edge (AE)}: $\sum_{ j \in \mathcal{K}(i) } {\mathit{gain}(i,j)} / |\mathcal{K}(i)| $; \textit{AECS}: $\sum_{ j \in \mathcal{K}(i) } {\frac{\mathit{gain}(i,j)}{c(i,j)}} / |\mathcal{K}(i)|$.
\end{itemize}

\renewcommand{\algorithmiccomment}[1]{\hfill~//#1}

\begin{algorithm}[tb]
   \caption{Spatial Graph UCT (SG-UCT).}
   \label{alg:sg-uct}
\begin{algorithmic}[1]
   \STATE {\bfseries Input:} spatial graph $G_0=(V,E_0,f,w)$,\\ objective function $\mathcal{F}$, budget $b_0$, reduction policy $\phi$
   \STATE {\bfseries Output:} actions $A_0, \dots A_{T-1}$ 
   \STATE \textbf{for} {$i$ in $V$}: compute $\mathcal{K}(i)$
   \STATE compute $\Phi = \phi(G_0)$ %
   \STATE $t=0$, $b_0 = \tau * C(E_0)$, $S_0=(G_0, \varnothing, b_0)$ 
   \STATE $\Delta_{max}= -\infty$
   \STATE $bestActs=array()$, $pastActs = array()$
   \STATE \textbf{loop}
   \begin{ALC@g}
      \STATE \textbf{if} $|\mathcal{A}_{\phi}(S_t)|=0$ \textbf{then} return $bestActs$
      \STATE create root node $v_t$ from $S_t$
      
      \STATE \textbf{for} {$i=0$ to $n_{sims}$}
      \begin{ALC@g}

         \STATE $v_l, treeActs =$ \textsc{TreePolicy}($v_t$, $\Phi$)
         \STATE $\Delta, outActs =$ \textsc{MinCostPolicy}($v_l$, $\Phi$) %
         \STATE \textsc{Backup}($v_l,\Delta$)
         \STATE \textbf{if} {$\Delta > \Delta_{max}$} \textbf{then} 
            \STATE \hskip1.0em $bestActs = [pastActs, treeActs, outActs]$ 
            \STATE \hskip1.0em $\Delta_{max} = \Delta$ %
      \end{ALC@g}

      \STATE $child=$ \textsc{MaxChild}($v_t$) 
      \STATE $pastActs.append(child.action)$
      \STATE $t+=1$, $S_t=child.state$
   \end{ALC@g}
\end{algorithmic}
\end{algorithm}

\section{Evaluation}\label{experiments}

\subsection{Experimental Protocol}\label{protocol}

\noindent \textbf{Definitions of Space and Distance.} For all experiments in this paper, we consider the unit 2D square as our space, i.e. we let $M=[0,1] \times [0,1]$ and the distance $d$ be Euclidean distance. In case the graph is defined on a spherical coordinate system (as is the case with physical networks positioned on Earth), we use the WGS84 variant of the Mercator projection to project nodes to the plane; then normalize to the unit plane. For simplicity, we consider uniform weights, i.e., $w(i,j)=1\ \forall\ (i,j) \in E$. 

\noindent \textbf{Synthetic and Real-World Graphs.} As a means of generating synthetic graph data, we use the popular model proposed by~\citet{kaiserSpatialGrowthRealworld2004}, which simulates a process of growth for spatial networks. Related to the Waxman model~\cite{waxmanRoutingMultipointConnections1988}, in this model the probability that a connection is created is inversely proportional to its distance from existing nodes. The distinguishing feature of this model is that, unlike e.g. the Random Geometric Graph~\cite{dallRandomGeometricGraphs2002}, this model produces connected networks: a crucial characteristic for the types of objectives we consider. We henceforth refer to this model as Kaiser-Hilgetag (shortened KH). We use $\alpha_{KH} = 10$ and $\beta_{KH}=10^{-3}$, which yields sparse graphs with scale-free degree distributions -- a structure similar to road infrastructure networks. We also evaluate performance on 7 networks belonging to the following real-world datasets: \textit{Internet} (a dataset of internet backbone infrastructure from a variety of ISPs~\cite{knightInternetTopologyZoo2011}) and \textit{Metro} (a dataset of metro networks in major cities around the world~\cite{rothLongtimeLimitWorld2012}). Due to computational budget constraints, we limit the sizes of networks considered to $|V|=150$.

\noindent \textbf{Setup.} For all experiments, we allow agents a modification budget equal to a proportion $\tau$ of the total cost of the edges of the original graph, i.e., $b_0 = \tau * C(E_0)$. We use $\tau = 0.1$. We let $\rho=1$ for synthetic graphs and $\rho=2$ for real-world graphs, respectively. Confidence intervals are computed using results of $10$ runs, each initialized using a different random seed. Rollouts are not truncated. We allow a number of node expansions per move $n_{sims}$ equal to $20 * |V|$ (a larger number of expansions can improve performance, but leads to diminishing returns), and select as the move at each step the node with the maximum average value (commonly referred to as \textsc{MaxChild}). The hyperparameter selection methodology and values are provided in the Technical Appendix.

\noindent \textbf{Baselines.} The baselines we compare against are detailed below. These methods represent the major approaches that have been considered in the past for the problem of goal-directed graph modifications: namely, evaluating a local node statistic~\cite{beygelzimer_improving_2005}, a shallow greedy criterion~\cite{schneider_mitigation_2011}, or the spectral properties of the graph~\cite{wangAlgebraicConnectivityOptimization2008,wang_improving_2014}. We do not consider previous RL-based methods since they are unsuitable for the scale of the largest graphs taken into consideration, for which the time needed to evaluate the objective functions makes training prohibitively expensive.

\begin{itemize}[leftmargin=*]
\item \textit{Random} ($\mathcal{F}_E, \mathcal{F}_R$): Randomly selects an available action.

\item \textit{Greedy} ($\mathcal{F}_E, \mathcal{F}_R$): Selects the edge that gives the biggest improvement in $\mathcal{F}$: formally, edge $(i,j)$ that satisfies $\argmax_{i,j} {\mathit{gain}(i,j)}$. We also consider the cost-sensitive variant \textit{Greedy\textsubscript{CS}}, for which the gain is offset by the cost: $\argmax_{i,j} {\frac{\mathit{gain}(i,j)}{c(i,j)}}$.

\item \textit{MinCost} ($\mathcal{F}_E, \mathcal{F}_R$): Selects edge $(i,j)$ that satisfies $\argmin_{i,j} {c(i,j)}$.

\item \textit{LBHB} ($\mathcal{F}_E$): Adds an edge between the node with Lowest Betweenness and the node with Highest Betweeness; formally, letting the betweeness centrality of node $i$ be $g_i$, this strategy adds an edge between nodes $\argmin_{i} g_i$ and $\argmax_{j} g_j$. 

\item \textit{LDP} ($\mathcal{F}_R$): Adds an edge between the vertices with the \textit{Lowest Degree Product}, i.e., vertices $i,j$ that satisfy $\argmin_{i,j} {d_i \cdot d_j}$.

\item \textit{FV} ($\mathcal{F}_R$): Adds an edge between the vertices $i,j$ that satisfy $\argmax_{i,j} {|\mathbf{y}_i - \mathbf{y}_j|}$, where $\mathbf{y}$ is the Fiedler Vector~\cite{fiedler1973algebraic,wangAlgebraicConnectivityOptimization2008}.

\item \textit{ERes} ($\mathcal{F}_R$): Adds an edge between vertices with the highest pairwise Effective Resistance, i.e., nodes $i,j$ that satisfy $\argmax_{i,j} {\Omega_{i,j}}$. $\Omega_{i,j}$ is defined as $(\hat{\mathcal{L}}^{-1})_{ii} + (\hat{\mathcal{L}}^{-1})_{jj} -2(\hat{\mathcal{L}}^{-1})_{ij}$, where $\hat{\mathcal{L}}^{-1}$ is the pseudoinverse of the graph Laplacian $\mathcal{L}$~\cite{wang_improving_2014}.

\end{itemize}

\subsection{Evaluation Results}

\begin{figure}
\begin{center}
\includegraphics[width=0.95\columnwidth]{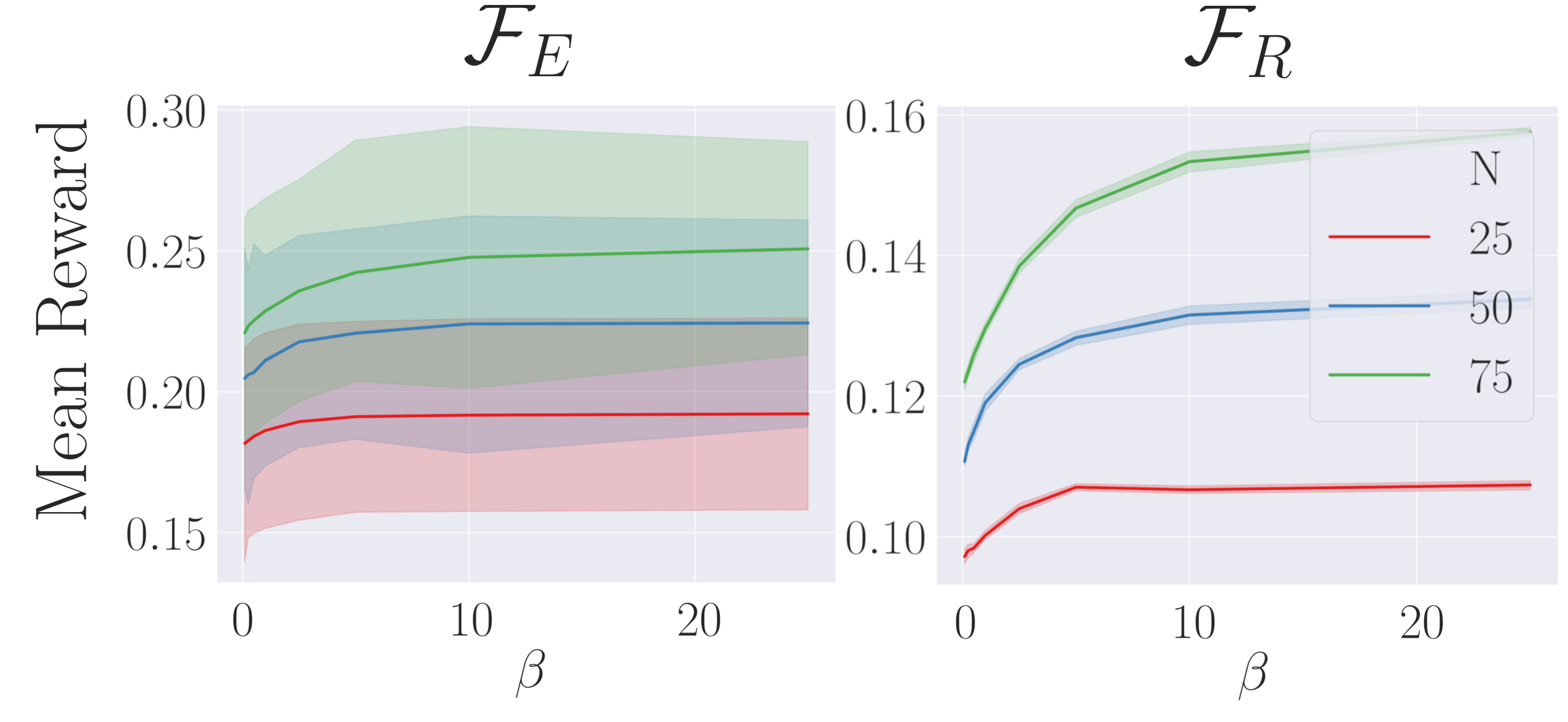}
\caption{Mean reward for SG-UCT\textsubscript{MINCOST} as a function of $\beta$, suggesting a bias towards low-cost edges is beneficial.}
\label{mincost_beta}
\end{center}
\end{figure}

\noindent \textbf{Synthetic and Real-world Graphs.} We consider $50$ synthetic KH graphs each of sizes $\{ 25, 50, 75 \}$ and the real-world graphs described previously. The obtained results are shown in Table~\ref{prelim-results}. On synthetic graphs, we find that SG-UCT outperforms UCT and all other methods in all the settings tested, obtaining 13\% and 32\% better performance than UCT on the largest synthetic graphs for the efficiency and robustness measures respectively. For $\mathcal{F}_{R}$, UCT outperforms all baselines, while for $\mathcal{F}_{E}$ the performance of the Greedy baselines is superior to UCT. Interestingly, MinCost yields solutions that are superior to all other heuristics and comparable to search-based methods while being very cheap to evaluate. Furthermore, UCT performance decays in comparison to the baselines as the size of the graph increases. On real-world graphs, we find that SG-UCT performs better than UCT and all other methods in all settings tested. The differences in performance between SG-UCT and UCT are 10\% and 39\% for $\mathcal{F}_E$ and $\mathcal{F}_R$ respectively. We note that the Greedy approaches do not scale to the larger real-world graphs due to their complexity: $O(|V|)^2$ choices need to be considered at each step, in comparison to the $O(|V|)$ required by UCT and SG-UCT. An extended version of the real-world graph results, split by individual graph instance, is shown in Table 3 in the Technical Appendix.

\begin{table*}[t]
\caption{Results obtained by the methods on synthetic and real-world graphs.}
\vskip 0.1cm
\label{prelim-results}
\begin{center}
\begin{small}
\begin{sc}
\resizebox{0.99\textwidth}{!}{
\begin{tabular}{cccccccccccc}
$\mathcal{F}$ & $\mathbf{G}$ &       Random &   LDP &    FV &  ERes &  MinCost &  LBHB & Greedy & Greedy\textsubscript{CS} &          UCT &       SG-UCT (ours) \\
\midrule

$\mathcal{F}_{E}$ 	& KH-25 & 0.128\tiny{$\pm0.008$} &   --- &   --- &   --- &    0.270 & 0.119 &  0.298 & 0.281 & 0.288\tiny{$\pm0.003$} &  \textbf{0.305\tiny{$\pm0.000$}} \\

					& KH-50 &  0.089\tiny{$\pm0.005$} &   --- &   --- &   --- &    0.303 & 0.081 & 0.335 & 0.311 & 0.307\tiny{$\pm0.003$} &  \textbf{0.341\tiny{$\pm0.000$}} \\

					& KH-75 &  0.077\tiny{$\pm0.004$}  &   --- &   --- &   --- &    0.315 & 0.072 & 0.339 & 0.319 & 0.311\tiny{$\pm0.003$} &  \textbf{0.352\tiny{$\pm0.001$}} \\

					& Internet &  0.036\tiny{$\pm0.005$} &   --- &   --- &   --- &    0.096 & 0.039 & $\infty$ & $\infty$ & 0.137\tiny{$\pm0.002$} &  \textbf{0.145\tiny{$\pm0.006$}} \\

					& Metro &  0.013\tiny{$\pm0.002$} &   --- &   --- &   --- &    0.049 & 0.007 & $\infty$ & $\infty$ &   0.056\tiny{$\pm0.001$} &  \textbf{0.064\tiny{$\pm0.000$}} \\

\midrule

$\mathcal{F}_{R}$ 	& KH-25 &  0.031\tiny{$\pm0.002$} &   0.049 &   0.051 &   0.054 &    0.065 & --- &  0.064 & 0.083 & 0.092\tiny{$\pm0.001$} & \textbf{0.107\tiny{$\pm0.001$}} \\

					& KH-50 &  0.033\tiny{$\pm0.002$} &   0.044 &   0.049 &   0.057 &    0.082 & --- & 0.078  &  0.102 & 0.112\tiny{$\pm0.002$} &  \textbf{0.140\tiny{$\pm0.001$}} \\

					& KH-75 &  0.035\tiny{$\pm0.002$} &   0.040 &   0.049 &   0.052 &    0.099 & --- & 0.074 & 0.115 &  0.120\tiny{$\pm0.001$} & \textbf{0.158\tiny{$\pm0.000$}} \\

					& Internet &  0.014\tiny{$\pm0.005$} & 0.025 & 0.015 & 0.021 &    0.072 &   --- & $\infty$ & $\infty$ &   0.083\tiny{$\pm0.002$} &  \textbf{0.128\tiny{$\pm0.006$}} \\

					& Metro &  0.009\tiny{$\pm0.002$} & 0.012 & 0.013 & 0.020 &    0.048 &   --- & $\infty$ & $\infty$ & 0.068\tiny{$\pm0.002$} &  \textbf{0.085\tiny{$\pm0.001$}} \\
\bottomrule
\end{tabular}

}
\end{sc}
\end{small}
\end{center}
\end{table*}

\noindent \textbf{Ablation Study.} We conduct an ablation study on synthetic graphs in order to assess the impact of the components of SG-UCT. The obtained results are shown in Table~\ref{ablation-results}, where SG-UCT\textsubscript{BTM} denotes UCT with best trajectory memorization, SG-UCT\textsubscript{MINCOST} denotes UCT with the cost-based default policy, SG-UCT\textsubscript{$\phi-q$} for $q$ in $\{40, 60, 80\}$ denotes UCT with a particular reduction policy $\phi$, and $q$ represents the percentage of original nodes that are selected by $\phi$. We find that BTM indeed brings a net improvement in performance: on average, 5\% for $\mathcal{F}_E$ and 11\% for $\mathcal{F}_R$. The benefit of the cost-based default policy is substantial (especially for $\mathcal{F}_R$), ranging from 4\% on small graphs to 27\% on the largest graphs considered, and grows the higher the level of bias. This is further evidenced in Figure~\ref{mincost_beta}, which shows the average reward obtained as a function of $\beta$. In terms of reduction policies, even for a random selection of nodes, we find that the performance penalty paid is comparatively small: a 60\% reduction in actions translates to at most 15\% reduction in performance, and as little as 5\%; the impact of random action reduction becomes smaller as the size of the network grows. The best-performing reduction policies are those based on a node's gains, with BECS and AECS outperforming UCT with no action reduction. For the $\mathcal{F}_R$ objective, a poor choice of bias can be harmful: prioritizing nodes with high degrees leads to a 32\% reduction in performance compared to UCT, while a bias towards lower-degree nodes is beneficial.

\begin{table}[t]
\caption{Ablation study for the components of SG-UCT.}
\vskip 0.1cm
\label{ablation-results}
\begin{center}
\begin{small}
\begin{sc}
\resizebox{0.99\columnwidth}{!}{
   \begin{tabular}{lllll}
& $|V|$ &           25 &           50 &           75 \\
\midrule
$\mathcal{F}_{E}$ & UCT             &  0.288\tiny{$\pm0.003$} &  0.307\tiny{$\pm0.003$} &  0.311\tiny{$\pm0.003$} \\
& SG-UCT\textsubscript{BTM}         &  0.304\tiny{$\pm0.001$} &  0.324\tiny{$\pm0.002$} &  0.324\tiny{$\pm0.002$} \\
& SG-UCT\textsubscript{MINCOST}     &  0.299\tiny{$\pm0.001$} &  0.327\tiny{$\pm0.001$} &  0.333\tiny{$\pm0.001$} \\
& SG-UCT\textsubscript{RAND-80}     &  0.284\tiny{$\pm0.005$} &  0.305\tiny{$\pm0.003$} &  0.303\tiny{$\pm0.003$} \\
& SG-UCT\textsubscript{RAND-60}     &  0.271\tiny{$\pm0.007$} &  0.288\tiny{$\pm0.004$} &  0.288\tiny{$\pm0.003$} \\
& SG-UCT\textsubscript{RAND-40}     &  0.238\tiny{$\pm0.009$} &  0.263\tiny{$\pm0.005$} &  0.271\tiny{$\pm0.003$} \\
& SG-UCT\textsubscript{DEG-40}      &  0.237\tiny{$\pm0.003$} &  0.262\tiny{$\pm0.003$} &  0.255\tiny{$\pm0.002$} \\
& SG-UCT\textsubscript{INVDEG-40}   &  0.235\tiny{$\pm0.002$} &  0.268\tiny{$\pm0.001$} &  0.283\tiny{$\pm0.001$} \\
& SG-UCT\textsubscript{NC-40} 		&  0.234\tiny{$\pm0.003$} &  0.268\tiny{$\pm0.002$} &  0.262\tiny{$\pm0.003$} \\
& SG-UCT\textsubscript{BE-40}       &  0.286\tiny{$\pm0.002$} &  0.304\tiny{$\pm0.002$} &  0.297\tiny{$\pm0.001$} \\
& SG-UCT\textsubscript{BECS-40}     &  0.290\tiny{$\pm0.001$} &  0.316\tiny{$\pm0.001$} &  0.319\tiny{$\pm0.002$} \\
& SG-UCT\textsubscript{AE-40}       &  0.286\tiny{$\pm0.002$} &  0.302\tiny{$\pm0.003$} &  0.297\tiny{$\pm0.002$} \\
& SG-UCT\textsubscript{AECS-40}     &  0.289\tiny{$\pm0.001$} &  0.317\tiny{$\pm0.001$} &  0.319\tiny{$\pm0.002$} \\
\midrule
$\mathcal{F}_{R}$ & UCT             &  0.092\tiny{$\pm0.001$} &  0.112\tiny{$\pm0.002$} &  0.120\tiny{$\pm0.001$} \\
& SG-UCT\textsubscript{BTM}         &  0.106\tiny{$\pm0.001$} &  0.123\tiny{$\pm0.001$} &  0.128\tiny{$\pm0.001$} \\
& SG-UCT\textsubscript{MINCOST}     &  0.105\tiny{$\pm0.001$} &  0.131\tiny{$\pm0.001$} &  0.153\tiny{$\pm0.001$} \\
& SG-UCT\textsubscript{RAND-80}     &  0.091\tiny{$\pm0.001$} &  0.111\tiny{$\pm0.001$} &  0.119\tiny{$\pm0.001$} \\
& SG-UCT\textsubscript{RAND-60}     &  0.089\tiny{$\pm0.003$} &  0.107\tiny{$\pm0.002$} &  0.115\tiny{$\pm0.002$} \\
& SG-UCT\textsubscript{RAND-40}     & 0.083\tiny{$\pm0.001$} &  0.102\tiny{$\pm0.002$} &  0.110\tiny{$\pm0.002$} \\
& SG-UCT\textsubscript{DEG-40}      &  0.069\tiny{$\pm0.001$} &  0.086\tiny{$\pm0.001$} &  0.092\tiny{$\pm0.001$} \\
& SG-UCT\textsubscript{INVDEG-40}   &  0.094\tiny{$\pm0.001$} &  0.114\tiny{$\pm0.001$} &  0.124\tiny{$\pm0.001$} \\
& SG-UCT\textsubscript{NC-40} 		&  0.071\tiny{$\pm0.001$} &  0.087\tiny{$\pm0.002$} &  0.092\tiny{$\pm0.001$} \\
& SG-UCT\textsubscript{BE-40}       &  0.088\tiny{$\pm0.001$} &  0.108\tiny{$\pm0.001$} &  0.115\tiny{$\pm0.001$} \\
& SG-UCT\textsubscript{BECS-40}     &  0.097\tiny{$\pm0.001$} &  0.115\tiny{$\pm0.001$} &  0.121\tiny{$\pm0.001$} \\
& SG-UCT\textsubscript{AE-40}       &  0.088\tiny{$\pm0.001$} &  0.103\tiny{$\pm0.001$} &  0.114\tiny{$\pm0.001$} \\
& SG-UCT\textsubscript{AECS-40}     &  0.098\tiny{$\pm0.001$} &  0.117\tiny{$\pm0.001$} &  0.126\tiny{$\pm0.001$} \\

\bottomrule
\end{tabular}

}
\end{sc}
\end{small}
\end{center}
\end{table}

\noindent \textbf{Runtime Analysis.} We also conduct a runtime analysis in order to compare the computational complexities of the considered methods. For this, we assume that the number of edges $|E|$ depends on the number of nodes $|V|$ by a constant factor $m$, i.e., $|E| \approx  m|V|$. This assumption is appropriate since the class of graphs under consideration is highly sparse (e.g., for the specific real-world networks considered, $m=1.191$). Determining the allowed actions and computing the cost of the edges is performed once at the start of execution, which costs $O(|V|)$ in the worst case via a KD-Tree. As discussed, for a generic network the robustness and efficiency objectives have complexity $O(|V|^2 \times (|V| + |E|))$ and $O(|V|^3)$ respectively. Given the above assumption regarding edge density, the complexity of the robustness objective becomes $O(|V|^2 \times (|V| + m|V|)) = O(|V|^2 \times ((m+1) |V|)) = O(|V|^2 \times |V|) = O(|V|^3)$. Thus, both objectives can be computed in $O(|V|^3)$. Then, the runtime complexity \textit{per edge addition} for the various methods is as follows:

\begin{itemize}
	\item Random: $O(1)$; LDP, LBHB: $O(|V|^2)$, since they involve computing the pairwise product of two arrays of size $|V|$;
	\item FV, ERes: $O(|V|^3)$, since they involve computing the (pseudo)inverse of the graph Laplacian;
	\item Greedy, GreedyCS: $O(|V|^5)$, since $\mathcal{F}$ must be computed in $O(|V|^3)$ after adding each of $O(|V|^2)$ possible edges;
	\item UCT, SG-UCT: $O(|V|^4)$, since the $O(|V|^3)$ computation of $\mathcal{F}$ is performed for $O(|V|)$ simulations.
\end{itemize}

Thus, our method also yields improved scalability compared to a standard greedy search, which was also observed empirically. While for certain network design problems (e.g., ~\citet{dilkina2011upgrading}) a greedy approach represents a reasonable approximation, it is too expensive to compute for this family of problems.

\subsection{Discussion}\label{discussion}

Our work is related to a large body of prior work in network design and optimization~\cite{ahujaChapterApplicationsNetwork1995}. Notably, the Upgrading Shortest Path~\cite{dilkina2011upgrading} and Pre-disaster Transportation Network Preparation~\cite{peeta2010pre,wu2016optimizing} problems resemble our case studies on efficiency and robustness improvement. Such works formulate the problem under study as a mathematical program on a case-by-case basis, for which known combinatorial solvers can be applied. In contrast, our method is generic and makes no assumptions about the objective at hand. In principle, any objective function defined on a spatial graph is admissible, allowing our method to optimize for complex, non-linear objectives that may arise in the real world (the resilience of the network to targeted attacks is one example of such a non-linear objective). Furthermore, our method can be applied for objectives for which no solutions are currently known without the expertise required to cast the problem as a mathematical program. Necessarily, the generic nature of our method means that it may perform worse on certain specific problems with linear constraints and objectives. The trade-offs arising from the use of generic machine learning and decision-making algorithms is a topic of ongoing debate in the combinatorial optimization community~\cite{bengio_machine_2018}, and we view these classes of methods as complementary.

\section{Conclusion}\label{conclusion}

In this work, we have addressed the problem of spatial graph construction: namely, given an initial spatial graph, a budget defined in terms of edge lengths, and a global objective, finding a set of edges to be added to the graph such that the value of the objective is maximized. For the first time among related works, we have formulated this task as a deterministic MDP that accounts for how the spatial geometry influences the connections and organizational principles of real-world networks. Building on the UCT framework, we have considered several concerns in the context of this problem space and proposed the Spatial Graph UCT (SG-UCT) algorithm to address them. 

Our evaluation results show performance substantially better than UCT (24\% on average and up to 54\% in terms of a robustness measure) and all existing baselines taken into consideration. Furthermore, our approach brings an important improvement in scalability with respect to recent work in deep RL for goal-directed graph construction~\cite{darvariu2021goal}, by scaling to graphs that are $3$ times larger in size and not requiring the expensive process of training a model for this task. We hope that our work will be used as a basis for the design of more cost-effective infrastructure systems such as communication, power, transportation, and water distribution networks.

\bibliography{bibliography}
\bibliographystyle{icml2022}

\newpage
\appendix
\onecolumn
\section{Reproducibility}\label{reproducibility}

\begin{wraptable}{R}{0.33\textwidth}
\vspace{-2\baselineskip}
\caption{Real-world graphs considered in the evaluation.}\label{tab:rw-graph-metadata}
\vskip 0.1cm
\resizebox{0.33\textwidth}{!}{
\begin{sc}
\begin{tabular}{cccc}
      Dataset & Graph &  $|V|$ &  $|E|$ \\
\midrule
 Internet &       Colt &        146 &        178 \\
          &      GtsCe &        130 &        169 \\
          &    TataNld &        141 &        187 \\
          &  UsCarrier &        138 &        161 \\
    Metro &  Barcelona &        135 &        159 \\
          &    Beijing &        126 &        139 \\
          &     Mexico &        147 &        164 \\
          &     Moscow &        134 &        156 \\
          &      Osaka &        107 &        122 \\
\bottomrule
\end{tabular}

\end{sc}
}
\vspace{-\baselineskip}
\end{wraptable}

\noindent \textbf{Implementation.} We implement all approaches and baselines in Python using a variety of numerical and scientific computing packages~\cite{hunter2007mpl,hagberg_exploring_2008,mckinney2011pandas,paszke2019pytorch,waskom2021seaborn}, while the calculations of the objective functions (efficiency and robustness) are performed in a custom C++ module as they are the main speed bottleneck. In a future version, we will release the implementation as Docker containers together with instructions that enable reproducing (up to hardware differences) all the results reported in the paper, including tables and figures.

\noindent \textbf{Data Availability.} In terms of the real-world datasets, the \textit{Internet} dataset is publicly available without any restrictions and can be downloaded via the Internet Topology Zoo website, \url{http://www.topology-zoo.org/dataset.html}. The \textit{Metro} dataset was originally collected by~\citet{rothLongtimeLimitWorld2012}, and was licensed to us by the authors for the purposes of this work. A copy of the \textit{Metro} dataset can be obtained by others by contacting its original authors for licensing (see \url{https://www.quanturb.com/data}). The considered graphs are detailed in Table~\ref{tab:rw-graph-metadata}.

\section{Additional Evaluation Details}\label{details}

\begin{figure*}[ht]
\begin{center}
\includegraphics[width=0.9\textwidth]{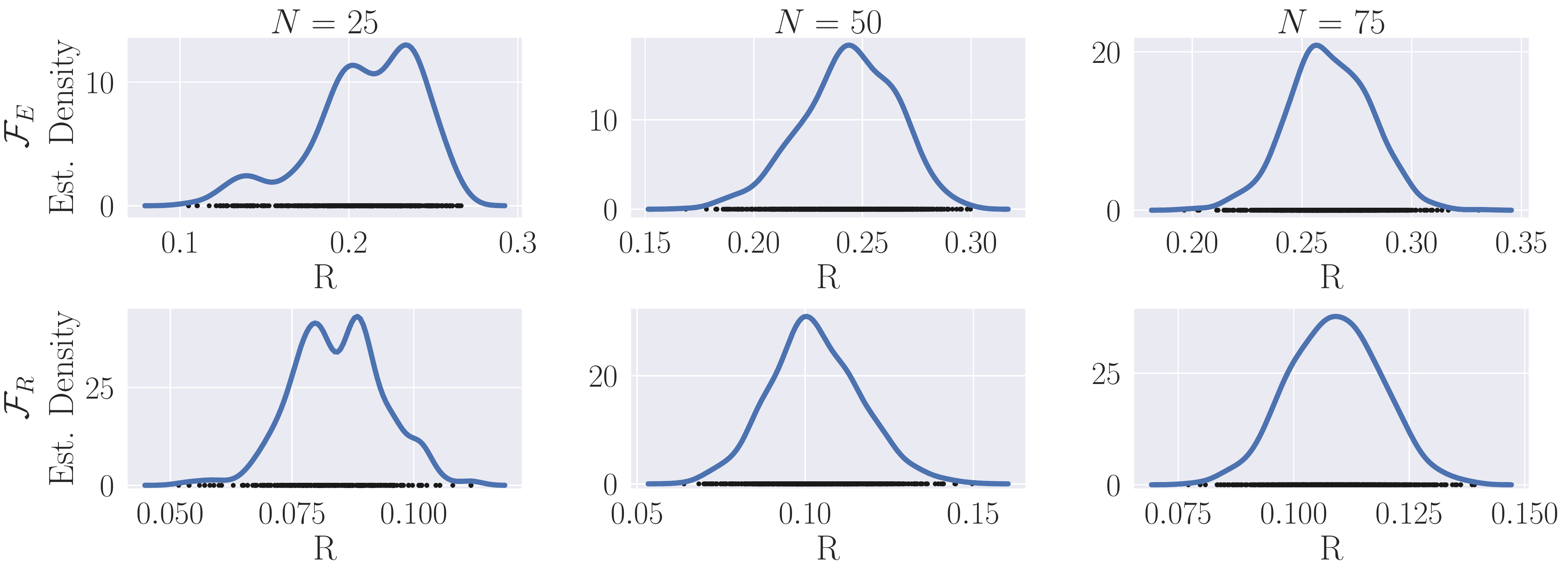} 
\caption{Empirical distribution of reward obtained for subsets selected by a uniform random $\phi$.}
\label{subsets_R_dist}
\end{center}
\end{figure*}

\textbf{Impact of Action Subsets on UCT Results.} We include an additional experiment related to the Action Space Reduction problem discussed in the Algorithm subsection of the manuscript. We consider, starting from the same initial graph, a selection of $1000$ subsets of size 40\% of all nodes, obtained with a uniform random $\phi$. We show the empirical distribution of the reward obtained by UCT with different sampled subsets in Figure ~\ref{subsets_R_dist}. Since the subset that is selected has an important impact on performance, a reduction policy yielding high-reward subsets is highly desirable (effectively, we want to bias subset selection towards the upper tail of the distribution of obtained rewards). 

\noindent \textbf{Hyperparameters.} Hyperparameter optimization for UCT and SG-UCT is performed separately for each objective function and synthetic graph model / real-world network dataset. For synthetic graphs, hyperparameters are tuned over a disjoint set of graphs. For real-world graphs, hyperparameters are optimized separately for each graph. We consider an exploration constant $c_p \in \{0.05, 0.1, 0.25, 0.5, 0.75, 1 \}$. Since the ranges of the rewards may vary in different settings, we further employ two means of standardization: during the tree search we instead use $\mathcal{F}(G_T)$ as the final reward $R_{T-1}$, and further standardize $c_p$ by multiplying with the average reward $R(s)$ observed at the root in the previous timestep -- ensuring consistent levels of exploration. The hyperparameters for the ablation study are bootstrapped from those of standard UCT, while $\beta \in \{0.1, 0.25, 0.5, 1, 2.5, 5, 10\}$ for the SG-UCT\textsubscript{MINCOST} variant is optimized separately. These results are used to reduce the hyperparameter search space for SG-UCT for both synthetic and real-world graphs. Values of hyperparameters used are shown in Table~\ref{tab-hyps}. For estimating $\mathcal{F}_R$ we use $|V|/4$ Monte Carlo simulations. 

\noindent \textbf{Extended Results.} Extended results for real-world graphs, split by graph instance, are shown in Table~\ref{rw-results-full}.

\noindent \textbf{Infrastructure and Runtimes.} Experiments were carried out on an internal cluster of 8 machines, each equipped with 2 Intel Xeon E5-2630 v3 processors and 128GB RAM. On this infrastructure, all experiments reported in this paper took approximately 21 days to complete.

\begin{table*}[h]
\caption{Hyperparameters used.}
\vskip 0.1cm
\label{tab-hyps}
\begin{center}
\begin{small}
\begin{sc}
\resizebox{0.7\textwidth}{!}{
\begin{tabular}{lllllllll}
         &     & {} & \multicolumn{2}{l}{$C_p$} & \multicolumn{2}{l}{$\phi$} & \multicolumn{2}{l}{$\beta$} \\
         &     & Objective & $\mathcal{F}_{E}$ & $\mathcal{F}_{R}$ &       $\mathcal{F}_{E}$ & $\mathcal{F}_{R}$ &      $\mathcal{F}_{E}$ & $\mathcal{F}_{R}$ \\
Experiment & Graph & Agent &            &              &                  &              &                 &              \\
\midrule
Internet & Colt & SG-UCT &       0.05 &         0.05 &      AECS-40 &  AECS-40 &              25 &           25 \\
         &     & UCT &        0.1 &          0.1 &              --- &          --- &             --- &          --- \\
         & GtsCe & SG-UCT &        0.1 &         0.05 &      AECS-40 &  AECS-40 &              25 &           25 \\
         &     & UCT &       0.25 &          0.1 &              --- &          --- &             --- &          --- \\
         & TataNld & SG-UCT &       0.05 &         0.05 &      AECS-40 &  AECS-40 &              25 &           25 \\
         &     & UCT &        0.1 &          0.1 &              --- &          --- &             --- &          --- \\
         & UsCarrier & SG-UCT &       0.05 &         0.05 &      AECS-40 &  AECS-40 &              25 &           25 \\
         &     & UCT &       0.05 &          0.1 &              --- &          --- &             --- &          --- \\
Metro & Barcelona & SG-UCT &       0.05 &         0.05 &      AECS-40 &  AECS-40 &              25 &           25 \\
         &     & UCT &       0.05 &         0.05 &              --- &          --- &             --- &          --- \\
         & Beijing & SG-UCT &       0.05 &         0.05 &      AECS-40 &  AECS-40 &              25 &           25 \\
         &     & UCT &       0.05 &         0.05 &              --- &          --- &             --- &          --- \\
         & Mexico & SG-UCT &       0.05 &         0.05 &      AECS-40 &  AECS-40 &              25 &           25 \\
         &     & UCT &       0.25 &         0.05 &              --- &          --- &             --- &          --- \\
         & Moscow & SG-UCT &       0.25 &          0.1 &      AECS-40 &  AECS-40 &              25 &           25 \\
         &     & UCT &       0.05 &         0.05 &              --- &          --- &             --- &          --- \\
         & Osaka & SG-UCT &       0.05 &         0.05 &      AECS-40 &  AECS-40 &              25 &           25 \\
         &     & UCT &       0.05 &         0.25 &              --- &          --- &             --- &          --- \\
KH-25 & --- & SG-UCT &       0.05 &         0.25 &      AECS-40 &  AECS-40 &              25 &           25 \\
         &     & UCT &        0.1 &          0.1 &              --- &          --- &             --- &          --- \\
         &     & SG-UCT\textsubscript{MINCOST} &        0.1 &          0.1 &              --- &          --- &              25 &           25 \\
KH-50 & --- & SG-UCT &       0.05 &         0.05 &      AECS-40 &  AECS-40 &              25 &           25 \\
         &     & UCT &       0.05 &         0.25 &              --- &          --- &             --- &          --- \\
         &     & SG-UCT\textsubscript{MINCOST} &       0.05 &         0.25 &              --- &          --- &              10 &           25 \\
KH-75 & --- & SG-UCT &       0.05 &         0.05 &      AECS-40 &  AECS-40 &              25 &           25 \\
         &     & UCT &       0.05 &          0.1 &              --- &          --- &             --- &          --- \\
         &     & SG-UCT\textsubscript{MINCOST} &       0.05 &          0.1 &              --- &          --- &              25 &           25 \\
\bottomrule
\end{tabular}

}
\end{sc}
\end{small}
\end{center}
\end{table*}

\begin{table*}[h]
\caption{Rewards obtained by baselines, UCT, and SG-UCT on real-world graphs, split by individual graph instance.}
\vskip 0.1cm
\label{rw-results-full}
\begin{center}
\begin{small}
\begin{sc}
\resizebox{0.9\textwidth}{!}{
\begin{tabular}{ccc|cccccccc}
             &       &  &       Random &   LDP &     FV &  ERes &  MinCost &  LBHB &          UCT &       SG-UCT (ours) \\
$\mathcal{F}$ & $\mathbf{G}$ & Graph &              &       &        &       &          &       &              &              \\
\midrule
$\mathcal{F}_{E}$ & Internet & Colt &  0.081\tiny{$\pm0.003$} &   --- &    --- &   --- &    0.127 & 0.098 &  0.164\tiny{$\pm0.003$} &    0.199\tiny{$\pm0.000$} \\
             &       & GtsCe &  0.017\tiny{$\pm0.007$} &   --- &    --- &   --- &    0.082 & 0.014 &  0.110\tiny{$\pm0.004$} &  0.125\tiny{$\pm0.002$} \\
             &       & TataNld &  0.020\tiny{$\pm0.004$} &   --- &    --- &   --- &    0.078 & 0.015 &  0.102\tiny{$\pm0.003$} &  0.110\tiny{$\pm0.002$} \\
             &       & UsCarrier &  0.026\tiny{$\pm0.014$} &   --- &    --- &   --- &    0.097 & 0.026 &  0.171\tiny{$\pm0.005$} &  0.178\tiny{$\pm0.002$} \\
             & Metro & Barcelona &  0.020\tiny{$\pm0.005$} &   --- &    --- &   --- &    0.063 & 0.003 &  0.067\tiny{$\pm0.002$} &  0.076\tiny{$\pm0.000$} \\
             &       & Beijing &  0.008\tiny{$\pm0.003$} &   --- &    --- &   --- &    0.028 & 0.003 &  0.041\tiny{$\pm0.001$} &  0.046\tiny{$\pm0.001$} \\
             &       & Mexico &  0.007\tiny{$\pm0.002$} &   --- &    --- &   --- &    0.032 & 0.011 &  0.037\tiny{$\pm0.001$} &  0.041\tiny{$\pm0.000$} \\
             &       & Moscow &  0.011\tiny{$\pm0.003$} &   --- &    --- &   --- &    0.038 & 0.007 &  0.043\tiny{$\pm0.001$} &  0.053\tiny{$\pm0.001$} \\
             &       & Osaka &  0.017\tiny{$\pm0.005$} &   --- &    --- &   --- &    0.082 & 0.010 &  0.093\tiny{$\pm0.003$} &  0.102\tiny{$\pm0.000$} \\
$\mathcal{F}_{R}$ & Internet & Colt &  0.007\tiny{$\pm0.004$} & 0.005 &  0.006 & 0.009 &    0.075 &   --- &  0.055\tiny{$\pm0.003$} &    0.089\tiny{$\pm0.000$} \\
             &       & GtsCe &  0.023\tiny{$\pm0.011$} & 0.048 &  0.017 & 0.031 &    0.099 &   --- &  0.098\tiny{$\pm0.005$} &  0.155\tiny{$\pm0.003$} \\
             &       & TataNld &  0.017\tiny{$\pm0.010$} & 0.011 & -0.002 & 0.013 &    0.074 &   --- &  0.093\tiny{$\pm0.006$} &  0.119\tiny{$\pm0.002$} \\
             &       & UsCarrier &  0.010\tiny{$\pm0.004$} & 0.035 &  0.038 & 0.033 &    0.041 &   --- &  0.085\tiny{$\pm0.007$} &  0.125\tiny{$\pm0.003$} \\
             & Metro & Barcelona &  0.020\tiny{$\pm0.007$} & 0.010 &  0.009 & 0.036 &    0.071 &   --- &  0.076\tiny{$\pm0.004$} &  0.115\tiny{$\pm0.002$} \\
             &       & Beijing &  0.004\tiny{$\pm0.004$} & 0.003 &  0.002 & 0.001 &    0.037 &   --- &  0.055\tiny{$\pm0.003$} &  0.062\tiny{$\pm0.001$} \\
             &       & Mexico &  0.007\tiny{$\pm0.004$} & 0.003 &  0.005 & 0.011 &    0.038 &   --- &  0.051\tiny{$\pm0.002$} &  0.068\tiny{$\pm0.001$} \\
             &       & Moscow &  0.013\tiny{$\pm0.005$} & 0.033 &  0.042 & 0.034 &    0.031 &   --- &  0.090\tiny{$\pm0.003$} &  0.109\tiny{$\pm0.002$} \\
             &       & Osaka &  0.003\tiny{$\pm0.006$} & 0.008 &  0.011 & 0.015 &    0.064 &   --- &  0.066\tiny{$\pm0.004$} &  0.072\tiny{$\pm0.001$} \\
\bottomrule
\end{tabular}

}
\end{sc}
\end{small}
\end{center}
\end{table*}

\end{document}